\documentclass[accepted]{article}

\usepackage{microtype}
\usepackage{graphicx}
\usepackage{booktabs} 
\usepackage{multirow}

\usepackage{amsmath}
\usepackage{amssymb}
\usepackage{bbm}
\usepackage{bm}
\usepackage{todonotes}
\usepackage{subfig}

\newcommand{\E}[2]{\mathbb E_{#1}\left[#2\right]}

\newcommand{\x}[0]{x}
\newcommand{\h}[0]{h}
\newcommand{\m}[0]{\mu}
\newcommand{\norm}[2]{||#1||_{#2}}

\usepackage{hyperref}

\usepackage{icml2019}

\icmltitlerunning{
  Generalized Negative Correlation Learning for Deep Ensembling
}

\begin{document}
\twocolumn[
    \icmltitle{
        Generalized Negative Correlation Learning for Deep Ensembling
    }




\begin{icmlauthorlist}
\icmlauthor{Sebastian Buschjäger}{tu}
\icmlauthor{Lukas Pfahler}{tu}
\icmlauthor{Katharina Morik}{tu}
\end{icmlauthorlist}

\icmlaffiliation{tu}{Artificial Intelligence Group, TU Dortmund University, Germany}

\icmlcorrespondingauthor{Sebastian Buschjäger}{sebastian.buschjaeger@tu-dortmund.de}

\icmlkeywords{Bias, Variance, Ensemble, Deep Learning}

\vskip 0.3in
]



\printAffiliationsAndNotice{}  

\begin{abstract}
Ensemble algorithms offer state of the art performance in many machine learning applications. A common explanation for their excellent performance is due to the bias-variance decomposition of the mean squared error which shows that the algorithm's error can be decomposed into its bias and variance. Both quantities are often opposed to each other and ensembles offer an effective way to manage them as they reduce the variance through a diverse set of base learners while keeping the bias low at the same time. 
Even though there have been numerous works on decomposing other loss functions, the exact mathematical connection is rarely exploited explicitly for ensembling, but merely used as a guiding principle. In this paper, we formulate a generalized bias-variance decomposition for arbitrary twice differentiable loss functions and study it in the context of Deep Learning. We use this decomposition to derive a Generalized Negative Correlation Learning (GNCL) algorithm which offers explicit control over the ensemble's diversity and smoothly interpolates between the two extremes of independent training and the joint training of the ensemble. We show how GNCL encapsulates many previous works and discuss under which circumstances training of an ensemble of Neural Networks might fail and what ensembling method should be favored depending on the choice of the individual networks. We make our code publicly available under \url{https://github.com/sbuschjaeger/gncl}.
\end{abstract}

\section{Introduction}
\label{sec:introduction}

Ensemble algorithms offer state of the art performance in many Machine Learning applications and often outperform single classifiers by a large margin. One of the main theoretical driving forces behind the understanding of ensembles is the bias-variance decomposition. The bias-variance decomposition decomposes the algorithm's error into two additive parts -- its bias and its variance. Hence, a good algorithm should try to minimize both at the same time which often leads to a difficult balancing act between the two quantities. Ensemble algorithms are well-known to reduce the variance if a diverse set of base models is trained while also keeping the bias low making them such an effective class of algorithms. 
The bias-variance decomposition has been mathematically proven for the mean-squared error, which sparked a plethora of different ensembling algorithms for different loss functions exploiting the general notion of `diversity' in ensemble construction \cite{Webb/2000,geurts/etal/2006,Brown/etal/2005,Melville/Mooney/2005,Lee/etal/2015,Zhou/Feng/2017a,Dvornik/etal/2019}.
Interestingly, even though there have been numerous works on decomposing other loss functions such as $0-1$ loss or exponential families (e.g. log-likelihood loss), none of these theoretical insights have directly inspired new learning algorithms. The general notion of diversity in an ensemble is still one of the main driving forces in designing new ensembling algorithms, while the exact mathematical connection is rarely exploited explicitly. 

We argue, that diversity can be hurtful sometimes and must be controlled with respect to the base learners. To do so, we formulate a generalized bias-variance decomposition and study it in the context of Deep Learning. From this decomposition, we derive two different algorithmic extremes: Either, we train the entire ensemble jointly in an end-to-end fashion or we train each model completely independent from each other. We present a generalization of Negative Correlation Learning (GNCL) that smoothly interpolates between these two extremes and thus can capitalize on the entire spectrum of methods in-between. We show, how GNCL generalizes many existing ensembling techniques in a single framework and use it to explore under which circumstances training of an ensemble might fail and what ensembling methods should be favored depending on the choice of the individual networks. Our contributions are: 
\begin{itemize}
    \item \textbf{A Generalized Bias-Variance Decomposition:} We present the first bias-variance decomposition for arbitrary twice differentiable loss functions.
    \item \textbf{Generalized Negative Correlation Learning:} From this decomposition we derive Generalized Negative Correlation Learning (GNCL) and show how it generalizes existing NCL-like algorithms into a single framework.
    \item \textbf{Experimental evaluation:} We compare our approach against state-of-the-art ensemble algorithms for Deep Learning methods. We show how GNCL smoothly interpolates between different ensembling techniques offering the overall best performance. Our code is available under \url{https://github.com/sbuschjaeger/gncl}.
    \item \textbf{Explanation of results:} Our theoretical results accurately explain when certain ensembling methods should be favored over others: For small capacity Neural Networks, End-to-End learning should be favored, whereas, for larger capacity models, ensembling should shift towards independent training of the individual models. 
\end{itemize}

The paper is organized as follows: The next section surveys related work and focuses on the bias-variance decomposition as well as ensembling methods in the realm of Deep Learning. Section \ref{sec:bias-variance} then derives the bias-variance decomposition, whereas section \ref{sec:gncl} formalizes it into the Generalized Negative Correlation Learning algorithm. In section \ref{sec:experiments} we experimentally evaluate our method and section \ref{sec:conclusion} concludes the paper.

\section{Related Work}
\label{sec:related-work}

The first bias-variance decomposition was proposed by Harry Markowitz in \cite{Markowitz/1952} for the mean squared error (MSE), which was later found to be one of the cornerstones of modern financial portfolio theory. Its first appearance - also for the MSE - in the Machine Learning community was due to Geman et al. in \cite{geman/etal/1992} which then sparked a series of different decompositions (see e.g. \cite{ueda/nakano/1996,domingos/2000,James/2003} and references therein). Most notable is the work by Domingos in \cite{domingos/2000} as it provides a set of consistent definitions for bias and variance and gives rise to a decomposition of the $0-1$ loss which fits the previous decompositions of the MSE. However, we note that these decompositions either focus on the mean squared error or the $0-1$ loss for binary classification problems, but not for general loss functions. James claims in \cite{James/2003} to give a generalized bias-variance decomposition for any symmetric loss functions by providing a set of different definitions of Bias and Variance. He then continues to choose those definitions which `fit' the original MSE decomposition best but never proves the consistency of these definitions. Moreover, as he notes, these definitions are only applicable for binary classification problems and not applicable for real-valued predictions. A similar decomposition has also been proposed in the context of Product Of Expert ensembles called the ambiguity decomposition. This decomposition also first appeared for the MSE and is equal to the bias-variance decomposition although derived from a distributional point of view \cite{Krogh/Vedelsby/1995, Heskes/1998}. Later, Hansen and Heskes give in \cite{Hansen/etal/2000} a generalized ambiguity decomposition for exponential families. The authors assume that an additive decomposition with two summands of the form $\text{Error} = \text{Bias} + \text{Variance}$ exist and then continue to show that an exponential family will always result in such a decomposition. They do not discuss distributions which do not fit this assumption. Most closely related to our approach is the work due to Jiang et al. in \cite{Jiang/etal/2017}. Here, the authors derive a Generalized Ambiguity Decomposition for twice differentiable loss functions. Similar to our approach, the authors also use a second-order Taylor approximation around the ensemble's prediction $f(x)$ but seem to ignore the remainder in their construction. Their paper focuses on binary classification losses with a single output and does not directly translate into a new learning algorithm. Our approach on the other hand also encapsulates multi-class problems and therefore is a natural generalization of previous work. Moreover, we present a novel learning objective and show how this objective encapsulates many well-known existing objectives presented in the literature.

Even though the exact theoretical connection between Bias and Variance for other loss functions was missing, the decomposition for the MSE sparked a multitude of different algorithms. In this paper, we focus on ensembles of Deep Nets and Neural Networks. However, we note that our generalized bias-variance decomposition does not assume any specific base learners, but is equally applicable to any base learner, e.g. Decision Trees. 
In the realm of Neural Networks, Negative Correlation Learning (NCL) is a direct application of the bias-variance decomposition and was first proposed by Liu et al. in \cite{Liu/Yao/1999} and later refined by Brown et al. in \cite{Brown/etal/2005}. 
Opitz et al. use NCL as inspiration to enforce diversity among neural networks in an ensemble by employing the cross-entropy loss between the individual experts' outputs \cite{Opitz/etal/2017}. Dvornik et al. propose in \cite{Dvornik/etal/2019} a similar ensembling technique but are more freely in their choices to enforce diversity. Specifically, they train each network on the cross-entropy loss but employ the cosine-similarity and KL-Divergence as a regularization term to enforce diversity. Webb et al. recently proposed a similar objective which they justify by viewing the ensemble as a product of experts. This leads to the minimization of the KL-Divergence to preserve most of the contribution of each expert with a coupling term similar to NCL \cite{Webb/etal/2019, Webb/etal/2020}. 
Lee et al. train in \cite{Lee/etal/2015, Lee/etal/2016} a diverse ensemble of classifiers by using Stochastic Multiple Choice Learning (SMCL). Instead of training all ensemble members on all the available data, they only update that member with the smallest loss. This way, the diversity which naturally occurs due to the random initialization, is promoted. 
Bagging \cite{Breiman/1996} has also been applied in the context of Deep Learning \cite{Brown/etal/2005, Zeiler/Fergus/2014, Lee/etal/2015, Lakshminarayanan/etal/2017, Zhu/etal/2019, Ovadia/etal/2019, Webb/etal/2019, Webb/etal/2020} which enforces diversity by training each expert individually on a bootstrap sample of the training data\footnote{To the best of our knowledge, there is no publication on feature bagging for training Deep Networks.}. Some works argue, that the random initialization of Deep Nets combined with stochastic gradient descent promotes enough diversity \cite{Lee/etal/2015, Lakshminarayanan/etal/2017, Ovadia/etal/2019, Devlin/etal/2019} so that bootstrap samples are not required. It is also noteworthy, that this training method sometimes occurs as a special case for certain hyperparameter settings \cite{Brown/etal/2005, Webb/etal/2019, Webb/etal/2020}, including this work. 
Joint training of the entire ensemble in an End-to-End fashion has also been proposed \cite{Brown/etal/2005, Webb/etal/2019, Webb/etal/2020, Opitz/etal/2017, Dutt/etal/2017, Lee/etal/2015}. This approach ignores the bias and variance of the individual experts but focuses on the ensemble's joint loss. Here, the literature is slightly more fragmented. End-To-End training also occurs in \cite{Brown/etal/2005, Opitz/etal/2017, Webb/etal/2019, Webb/etal/2020} as a special case for certain hyperparameter settings, including this work. Dutt et al. call this approach a `coupled ensemble' \cite{Dutt/etal/2017}, whereas Lee et al. call this approach training under an `ensemble-aware' loss \cite{Lee/etal/2015}. 
While less extensive, Boosting \cite{Schapire/Freund/2012} has also been applied to Neural Networks and Deep Learning. Early works focused on the combination of smaller Neural Networks as base learners for ensembling which also carried over to larger architectures commonly found in Deep Learning \cite{Opitz/Maclin/1997, maclin/Opitz/1997, Schwenk/Bengio/2000, Moghimi/etal/2016, Zhu/etal/2019}. For reference, we note that there has also been some interest in understanding residual architectures (ResNet) as boosting in feature space \cite{Huang/etal/2018}.
Recently, ensembles which a derived from a single network have also been proposed. Dropout \cite{Srivastava/etal/2014} is anecdotally sometimes referred to as `the ensemble of possible subnetworks' \cite{Baldi/Sadowski/2013, Gal/etal/2016}. This connection has been studied more closely in the context of `pseudo-ensembles' \cite{Bachman/etal/2014}. Pseudo-ensembles are ensembles that are derived from a large single network by perturbing it with a noise process, e.g. by removing weights as done by Dropout. Although not explicitly mentioned, `snapshot ensembles' \cite{Qiu/etal/2014, Huang/etal/2017a} which store multiple versions of the same network during the optimization (e.g. by storing the current model every $10$ epochs) can also be seen in this framework. We will later revisit these methods and show how they relate to our GNCL approach.




\section{\resizebox{0.95\columnwidth}{!}{A Generalized Bias-Variance Decomposition}}
\label{sec:bias-variance}

We consider a supervised learning setting, in which we assume that training and test points are drawn i.i.d. according to some distribution $\mathcal D$ over the input space $\mathcal X$ and labels $\mathcal Y$. For training, we have given a labelled sample $\mathcal{S} = \{(x_i,y_i)|i=1,\dots,N\}$, where $x_i \in \mathcal X \subseteq \mathbb R^d$ is a $d$-dimensional feature-vector and $y_i\in \mathcal Y \subseteq \mathbb R^C$ is the corresponding target vector. For binary classification problems we set $C=1$ and $\mathcal Y = \{-1,+1\}$; for regression problems we have $C=1$ and $\mathcal Y = \mathbb R$. For multiclass problems with $C$ classes we encode each label as a one-hot vector $y = (0,\dots,0,1,0,\dots,0)$ which contains a `$1$' at coordinate $c$ for label $c \in \{0,\dots,C-1\}$. 

Given a model class $\mathcal H = \{\h \colon \mathcal X \to \mathbb R^C\}$ we wish to select that model that fits our current sample. In practice we employ various learning algorithms to do so, e.g. SGD in the context of Deep Learning or CART in the context of decision trees. These algorithms often introduce some form of randomization, e.g. by random initialization of weights or random sampling of splits thereby introducing some distribution $\Theta$ over possible models in $\mathcal H$ . Yet, in the heart of these algorithms, we find the minimization of a loss function $\ell\colon \mathcal Y \times \mathbb R^C \to \mathbb R_+$ which quantifies the error of our model's prediction $\h(x)$ compared to the real target $y$. So to choose the optimal algorithm for a problem, we may favor that algorithm which consistently produces the best models with the smallest loss: 
$$
\Theta^* = \arg\min_{\Theta}\E{h \sim \Theta,(x,y)\sim\mathcal D}{\ell(h(x),y)}
$$
For the rest of this paper we assume that $\ell$ is at least twice continuous differentiable and present appropriate choices at the end of this section. We now use a second-order Taylor approximation of $\ell$ around the centre $\m(x) = \E{h\sim \Theta}{h(x)}$. For readability we now drop the subscript $h \sim \Theta,(x,y)\sim\mathcal D$. Similarly we write $h(x) = h$ and $\m(x) = \m$ and $\ell(h(x),y) = \ell(h)$:
\begin{align*}
    \E{}{\ell(h)}
    &= 
    \E{}{\ell(\mu)} +
    \E{}{(\h - \m)^T\nabla_{\m} \ell(\m)} \\
    &\phantom{=}\, + \E{}{\frac{1}{2} (\h-\m)^T \nabla^2_{\m}\ell(\m) (\h-\m)} + \E{}{R_3} 
\end{align*}
where $R_3$ denotes the remainder of the Taylor approximation containing the third and higher derivatives.

We note, that $\nabla_{\m} \ell(\m)$ does not depend on $\h$ since $\m$ is a constant given a fixed test point $\x$ and therefore $\E{}{\nabla_{\m}  \ell(\m)} = \nabla_{\m} \ell(\m)$. Also note, that per definition $\E{}{h} = \m$ so that the second summand vanishes: 
\begin{align*}
\E{\h}{(\h - \m)^T\nabla_{\m} \ell(\m))} 
    &= 0
\end{align*}

Naturally, the quality of this approximation depends on the magnitude of the remainder and it becomes exact if the loss function does not have a third derivative. Otherwise, we may use a classic text-book (see e.g. \cite{edwards/2012, koenigsberger/2013} and the appendix for more details) result to bound the magnitude of the remainder for functions which are $3$ times continuous differentiable. Let there be some $m\in\mathbb R$ so the third derivative of the loss is bounded by it, that is $|\nabla^3_{\mu} \ell(\mu)|_{i,j,k} \le m$ for all $(x,y) \sim \mathcal D$ then
$$
R_3(\h - \m) \le \frac{1}{6}m \max_{\h}\norm{\h - \m|}{1}^3 \le \frac{1}{6} m C \max_{h_1,\dots,h_{C}} (h_i - \mu_i)^3
$$
For a sufficiently small remainder we approximate:
\begin{align}
    \label{eq:covar-decomposition}
    \E{}{\ell(h)} 
        &\approx \E{}{\ell(\mu)} + \E{}{\frac{1}{2} \phi^T \nabla^2_{\m}\ell(\m) \phi} \\ 
        &= \E{}{\ell(\mu)} + \frac{1}{2}tr\left(\nabla_{\m} \ell(\m) \texttt{cov}( \phi,  \phi)\right) 
\end{align}
where $\phi = (\h-\m)$ and the second line is the quadratic form of the expectation. We may interpret this decomposition as a generalized Bias-(Co-)Variance decomposition: While the LHS depicts the expected error of a model $h$, the first term on the RHS depicts the error of the expected model - or differently coined the \textit{algorithm's} bias. The second term can be interpreted as the co-variance of $h$ with respect to the expected model $\m$ given a loss-specific multiplicative constant $\nabla^2 \ell(\mu)$.  

\subsection{Example 1: Mean-squared error}
Consider the mean squared error (MSE) of a one dimensional regression task $\mathcal Y = \mathbb R$ and let $z = h(x)$:
\begin{align*}
    \ell(z,y) &= \frac{1}{2}(z - y)^2 \\
    \frac{\partial \ell}{\partial z} &= (z - y) \\
    \frac{\partial^2 \ell}{\partial z \partial h(x)} &= 1 \\
    \frac{\partial^3 \ell}{\partial z \partial h(x) \partial z} &= 0 
\end{align*}
The third derivative of the MSE vanishes and thus the above approximation is exact. The resulting decomposition matches exactly the well-known Bias-Co-Variance decomposition.

\subsection{Example 2: Negative-likelihood Loss}
As a second example we consider multi-class classification problem with $C$ classes. Let $z = h(x) \in \mathbb R^C$ and let $\ell$ be the negative-likelihood loss (NLL):
\begin{align*}
    \ell(z,y) &= -\sum_{i=1}^C y_i \log(z_i) \\
    \frac{\partial \ell}{\partial z_i} &= -\frac{y_i}{z_i} \\
    \frac{\partial^2 \ell}{\partial z_i \partial z_j} &= \frac{y_i}{z_i^2}\mathbbm 1\{i=j\} \\
    \frac{\partial^3 \ell}{\partial z_i \partial z_j \partial z_k} &= -2\frac{y_i}{z_i^3}\mathbbm 1\{i=j=k\} 
\end{align*}
For this loss function, the third derivative does not vanish and thus the decomposition is not exact. Looking at the third derivative we also see, that it can get uncontrollably large for $z_i \to 0$ if $y_i = 1$. Thus, if a model completely fails with a wrong prediction then the decomposition error can be unbounded. Put differently, the performance of a model using the NLLLoss cannot be completely explained in terms of `Bias' and `Variance' since the remainder is not neglectable. 

\subsection{Example 3: Cross Entropy Loss}
As a third example we consider the common combination of the NLLLoss with the softmax function, also called the Cross Entropy Loss. Again, let $z = h(x) \in \mathbb R^C$. The softmax function maps each output dimension $z_i$ of the classifier to a probability:
$$
q_i = \frac{e^{z_i}}{\sum_{i=1}^C e^{z_j}}
$$
We combine softmax with the NLLLoss:
\begin{align*}
    \ell(z,y) &= -\sum_{i=1}^C y_i \log(z_i) = -\sum_{i=1}^C y_i \log \left(\frac{e^{z_i}}{\sum_{i=1}^C e^{z_j}} \right) \\
    \frac{\partial \ell}{\partial z_i} &= q_i - \mathbbm 1\{y_i = 1\} \\
    \frac{\partial^2 \ell}{\partial z_i \partial z_j} &= q_i \left( \mathbbm 1\{i=j\} -q_j\right) \\
    \frac{\partial^3 \ell}{\partial z_i \partial z_j \partial z_k} &= \mathbbm 1\{i=j\}q_i \left( \mathbbm 1\{i=k\} - q_k \right) \\
    &\phantom{=} - q_i q_j\left( \mathbbm 1\{i=k\} - q_k \right) \\
    &\phantom{=} - q_i q_j \left( \mathbbm 1\{j=k\} - q_k \right) 
\end{align*}
Due to the softmax function we have $\sum_{c=1}^C q_c = 1, q_c > 0~\forall c = 1,\dots,C$. The maximum of the third derivative is obtained for pairwise unequal $i,j,k$ ($i \not= j, j \not= k, i \not= k$) and $q_i = q_j = q_k = \frac{1}{3}$:
$$
2 \cdot q_i q_j q_k \le \frac{1}{27} < 0.038
$$
Thus, the decomposition error for the cross entropy loss is bounded and we can explain a models performance in terms of its Bias and Variance (up to the bounded remainder).  


\section{Generalized Negative Correlation Learning}
\label{sec:gncl}

As often faced in Machine Learning we cannot compute $\E{x,y \sim \mathcal D}{\ell(h(x),y)}$ exactly since we do not know the exact distribution $\mathcal D$ and in fact, this is part of the problem we would like to solve. Moreover, it is difficult to compute $\m(x)=\E{h\sim\Theta}{h(x)}$ exactly since the algorithm we use for computing $h$ (e.g. SGD) only implicitly induces a distribution over $h$ and the exact nature of $\Theta$ for various model classes is ongoing research \cite{Biau/Scornet/2016, Sutskever/etal/2013, Arora/etal/2019, Kawaguchi/etal/2017}. For sufficiently large training sample $\mathcal S$ we use Monte-Carlo approximation: 
$$
\E{x,y}{\ell(h(x),y)} \approx \frac{1}{N}\sum_{(x,y)\in\mathcal S} \ell(h(x),y)
$$
Similarly, we may approximate the expected prediction $\mu$ with $M$ models:
\begin{align*}
    \mu(x) = \E{\Theta}{h(x)} &\approx f(x) = \frac{1}{M} \sum_{i=1}^M h^i(x) \\
    \E{\Theta}{\frac{1}{2} \phi^T \nabla^2_{\m}\ell(\m) \phi} &\approx \frac{1}{2M}\sum_{i=1}^M {d_i}^T D d_i  \\
    \E{\Theta}{R(x)} &\approx \widetilde R
\end{align*}
where $D = \nabla^2_{f(x)}\ell(f(x), y)$ and $d_i = (h^i(x) - f(x))$. We stress the fact, that we \textit{assume} that these are good approximations. For large $M$, this is certaintly a justified approximation, but for smaller $M$ this is not necessarily the case. However, additive ensembles of this form are arguably the most common form of ensembles and undeniably work well in practice.
We define the empirical bias-variance decomposition for any twice-differentiable loss function as:
\begin{equation}
    \label{eq:emperical-covar-decomposition}
    \ell(f) = \frac{1}{M} \sum_{i=1}^M \ell(h^i) - \frac{1}{2M} \sum_{i=1}^M {d_i}^T D d_i + \widetilde R
\end{equation}
We note, that for any convex loss function $D$ is positive definite and therefore ${d_i}^T D d_i \ge 0$ which implies:
\begin{equation*}
\ell(f) \le \frac{1}{M} \sum_{i=1}^M \ell(h^i)
\end{equation*}
It follows, that an ensemble of models will always be better than a single model making a compelling argument for ensemble learning. Note, that a similar argument has been made numerous times already and can for example directly be obtained when applying Jensen's inequality to the weighted average of models over a convex loss.

We use Eq. \ref{eq:emperical-covar-decomposition} as a basis for a learning algorithm: We can either directly minimize its LHS and optimize the entire ensemble in an end-to-end fashion. Alternatively, we use its RHS to derive a regularized objective which trains each network independently with a coupling term enforcing some diversity. To do so, let $\widetilde R$ be sufficiently small, so that we may ignore it and let $\lambda \in \mathbb R$ be a regularization parameter, then we may minimize:  
\begin{equation}
    \label{eq:GNCL2}
    \frac{1}{M} \sum_{i=1}^M \ell(h^i) - \frac{\lambda}{2M} \sum_{i=1}^M {d_i}^T D d_i
\end{equation}

\subsection{A combined loss function}
Having the two objectives $\ell(f)$ and eq. \ref{eq:GNCL2} available begs the question of which of both may lead to better results. Frankly, since both objectives are equal, minimizing both will lead to similar if not equal results. Thus, using either approach comes down to the more practical specifics of the problem at hand: Direct minimization of the loss seems favorable because it automatically finds a good trade-off between bias and variance and no hyperparameter tuning is necessary. Yet, using Eq. \ref{eq:GNCL2} on the other hand enables us to train each model independently and only requires some synchronization between models to make sure that the variance is large enough (See e.g. \cite{Webb/etal/2019} and references therein for a discussion on distributed training). Moreover, this approach allows practitioners to fine-tune the trade-off between bias and variance which might be favorable for specific problems and base models. 
In Deep Learning, it is common practice to train networks to achieve zero loss on the training data and sometimes train it even longer \cite{Zhang/etal/2017}. Recall that for a convex loss ${d_i}^T D d_i\ge 0$ and therefore Eq. \ref{eq:emperical-covar-decomposition} implies that (for a sufficiently small remainder) an ensemble with powerful base learners having zero training loss should not have any variance on the training data. Therefore, as soon as the base learners achieve zero training loss there is no need to invest into variance because the best model (from the training data's perspective) has already been found. Clearly, this is neither the intuition behind the bias-variance decomposition nor is it what we are trying to achieve. And indeed, in most practical applications we can be sure that even though we have zero training loss, that we will suffer some loss when applying our model to new, unseen data. In this case, it might still be favorable to enforce some diversity between base models during training to achieve a better generalization error. We will investigate this effect in our experiments more carefully and show that there is a clear dependence on which method to favor depending on the type and strength of the base learner.

Interestingly, there is an upper bound of the bias-variance decomposition that combines both approaches into a single objective. This upper bound simply re-scales the indivdual contributions of the base learners and thus results in the \textit{same} solution as minimizing $l(f)$ or Eq. \ref{eq:GNCL2} for appropriate choices of $\lambda$ which allows us to smoothly interpolate between the two extremes of independent and end-to-end training. In addition, this formulation circumvents the costly computation of $D$ and does not require the assumption that $\widetilde R$ is sufficiently small:
\begin{align*}
&\frac{1}{M} \sum_{i=1}^M \ell(h^i) - \frac{1}{2M} \sum_{i=1}^M {d_i}^T D d_i + \widetilde R \\ 
&\le \frac{1}{M} \sum_{i=1}^M \ell(h^i) - \frac{1}{2M} \sum_{i=1}^M {d_i}^T D d_i + \widetilde R  + \frac{1}{M}\sum_{i=1}^M \ell(h^i)\\
&= \ell(f) + \frac{1}{M}\sum_{i=1}^M \ell(h^i)
\end{align*}
To this end we propose the following Generalized Negative Correlation Learning (GNCL) objective for $\lambda \in [0,1]$:
\begin{equation}
    \label{eq:GNCL1}
    \frac{1}{N}\sum_{j=1}^N \left( \lambda \ell(f(x_j),y_j) + \frac{1-\lambda}{M}\sum_{i=1}^M \ell(h^i(x_j),y_j) \right)
\end{equation}
For $\lambda = 0$ this trains $M$ models independently, whereas for $\lambda = 1$ all models are trained jointly in an end-to-end fashion. For values between zero and one we can smoothly interpolate between these to extremes making the entire spectrum of trade-offs available. 


\subsection{Relationship to other ensembling approaches}
There are multiple mentions of NCL-like algorithms in literature. We will now show, that these algorithms are a special version of the proposed Generalized Negative Correlation Learning algorithm. 

\textit{Negative Correlation Learning:} The earliest works \cite{Liu/Yao/1999,Brown/etal/2005} on NCL-Learning propose to minimize the MSE with a coupling term including the ensembles' diveristy (c.f. Eq. (17) in \cite{Brown/etal/2005}): 
$$
\frac{1}{M}\sum_{i=1}^M \frac{1}{2} (h^i(x) - y)^2 - \lambda \frac{1}{M}\sum_{i=1}^M \frac{1}{2} (h^i(x) - f(x))^2
$$
Substituting the second derivative of the MSE loss in Eq. \ref{eq:GNCL2} directly leads to this formulation. NCL is a specialized version of GNCL for the MSE loss. 

\textit{Modular loss:} Webb et al. propose to minimize both, the ensemble loss as well as the loss of each individual expert in a modular loss function (c.f. Eq (4) in \cite{Webb/etal/2019}):
$$
\lambda \text{KL}(f(x) \| y) + \left( 1 - \lambda \right)\frac{1}{M}\sum_{i=1}^M \text{KL}(h^i(x) \| y)
$$
where $\text{KL}$ denotes the KL-Divergence and $\lambda \in [0,1]$ is the regularization strength. Substituting the cross-entropy loss into Eq. \ref{eq:GNCL1} yields the same formulation. The modular loss is a specialized version of GNCL with the cross entropy loss.

\textit{DivLoss:} Opitz et al. use NCL as inspiration to enforce diversity among neural networks by employing the cross-entropy loss between the individual experts' outputs while minimizing the individual and the ensemble loss. They propose to minimize (c.f. Eq. (15) in \cite{Opitz/etal/2017}) the DivLoss:
\begin{align*}
\ell(f(x), y) + &\frac{\lambda_1}{M}\sum_{i=1}^M \ell(h^i(x), y) \\
&~ - \frac{\lambda_2}{M (M-1)} \sum_{i=1}^M \sum_{j\not= i} \ell(h^i(x), h^j(x)) 
\end{align*}
where $\ell$ is the cross-entropy loss with softmax activation and $\lambda_1,\lambda_2 \in \mathbb R$ are regularization parameters. Note that $\ell(f(x), y) \ge 0$ and that $\ell(h^i(x), h^j(x))$ is convex in its first argument. For a fixed scaling $\kappa \le 1$ it holds that
\begin{align*}
\kappa \frac{1}{M}\sum_{i=1}^M \ell(h^i(x), y) &\le \ell \left( \frac{1}{M}\sum_{i=1}^M h^i(x), y\right)
\end{align*}
due to Jensen's inequality. Therefore, we lower-bound the original objective for $\lambda_2 \le 1$ to:
$$
\frac{\lambda_1}{M}\sum_{i=1}^M \ell(h^i(x), y) - \frac{\lambda_2}{(M-1)} \sum_{j=1}^M \ell \left(f(x), h^j(x) \right)
$$
Interestingly, Webb et al. show in \cite{Webb/etal/2019} that this formula is an  alternative fomulation of their modular loss when setting $\lambda_1 = 1$ and $\lambda_2 = \lambda \in [0,1]$. It follows, the objective proposed in \cite{Opitz/etal/2017} is an upper bound of the modular loss proposed in \cite{Webb/etal/2019}, which in turn is a specialized version of GNCL learning for the cross entropy loss. 

\textit{Diversity with Cooperation:} Dvornik et al. propose in \cite{Dvornik/etal/2019} an ensemble approach that focuses on diversity and cooperation at the same time. More formally, they propose to use the following objective
$$
\sum_{i=1}^M \ell(h^i(x), y) + \frac{\lambda}{(M-1)} \sum_{i=1}^M \sum_{j\not= i} \psi(h^i(x), h^j(x))
$$
where $\psi$ is a penalty function to enforce diversity in the ensemble. By using the cross entropy loss and setting $\psi = -\ell$ we arrive at the DivLoss function for $\lambda_1 = M$ and $\lambda_2 = 1$. Thus, the diversity with cooperation approach by Dvornik et al. is closely related to GNCL. However, we note that the authors are freer with their choices of $\psi$ leading to mixed experimental results.


\textit{Bagging and Wagging:} Bagging uses bootstrap samples to assign a different subset of training examples to each expert and thereby enforces diversity. Bauer and Kohavi \cite{Bauer/Kohavi/1999} propose an extension called Wagging which samples different weights instead of sampling examples directly. Oza and Russel show in \cite{Oza/Russel/2001} that Wagging with weights sampled from a discrete Poisson distribution $w \sim Poisson(1)$ is the same as Bagging. Similarly, Webb et al. propose in \cite{Webb/2000} to use continuous Poisson weights for Wagging, which improves the performance for certain base learners. Formally, the loss function for Wagging and its variants is
$$
\frac{1}{N}\sum_{j=1}^N \frac{1}{M}\sum_{i=1}^M w_{j,i} \widetilde \ell(h^i(x_j), y_j) 
$$
where $w_{j,i} \in \mathbb R_+$ is the precomputed weight for each base learner and sample and $\widetilde \ell \colon \mathcal Y \times \mathbb R^C \to \mathbb R_+$ is another loss function. Setting $\ell(h^i(x_j), y_j) = w_{i,j} \widetilde \ell(h^i(x_j), y_j)$ and $\lambda = 0$ in Eq. \ref{eq:GNCL1} yields the same formulation. Hence, we can simulate Wagging and Bagging with appropriate loss functions inside the GNCL framework.

\textit{Boosting:} Boosting iteratively trains new classifiers to correct the errors of the previous classifier, thereby constructing a strong classifier from weak base models. It is well-known that Boosting can be viewed as functional gradient descent in which each new base learner tries to approximate the negative gradient of a loss function \cite{Mason/etal/2000, Schapire/Freund/2012}. Thus, boosting minimizes the entire ensemble loss and is therefore closely related to GNCL with $\lambda = 1$. However, we note that Boosting fundamentally behaves differently from GNCL because it is designed to greedily approximate gradients for non-differential base learner's functions. Theoretically, both approaches could be combined: The proposed GNCL objective can either be minimized via Stochastic Gradient Descent (as done in our experiments) or by boosting weak-learner on the GNCL objective. 
Last, note that Residual Networks (ResNet) have also been shown to share a connection with Boosting. ResNets perform boosting in feature space with a telescoping sum instead of a weighted average as used by regular Boosting \cite{Huang/etal/2018} and thus -- while there is some overlap -- both methods are fundamentally different. Interestingly, we found that ResNet architectures performed best as a base learner during our experiments. 

\textit{Dropout and Pseudo-Ensembles:} Dropout \cite{Srivastava/etal/2014} is a regularization method for Deep Nets, which randomly sets weights to zero during the forward pass. While Dropout helps to prevent overfitting, it can also be used to estimate the geometric mean and variance of a distribution of networks with paramter sharing. This is anecdotally sometimes referred to as `the ensemble of possible subnetworks' \cite{Baldi/Sadowski/2013, Gal/etal/2016}. Bachman et al. studied this connection more closely and proposed in \cite{Bachman/etal/2014} the term `pseudo-ensembles'. Pseudo-ensembles are ensembles that are derived from a large single network by perturbing it with a noise process, e.g. by removing weights as done by Dropout. Although not explicitly mentioned, snapshot ensembles \cite{Qiu/etal/2014, Huang/etal/2017a} which store multiple versions of the same network (e.g. by storing the current model every $10$ epochs) can also be seen in this framework. Pseudo-ensembles minimize the following objective
$$
\frac{1}{N}\sum_{j=1}^N \E{\theta}{\ell_{\theta}(\mu(x_i),y_i)} + \lambda \E{\theta}{R\left(\mu(x_i), \mu_{\theta}(x_i)\right)}
$$
where $\mu$ denotes the `mother' net, $\mu_{\theta}$ is a child net under the noise process $\theta$, $\ell$ is a loss function and $R$ is a regularizer with regularization strength $\lambda$. Note, that for our bias-variance decomposition we derived the same objective in eq. \ref{eq:covar-decomposition} with $R = \phi^T \nabla^2_{\m}\ell(\m) \phi$ and by introducing $\lambda$ as discussed earlier. Unfortunatley, the authors do not discuss how to directly minimize this objective under the noise process $\theta$. Interestingly, for experiments they use the \textit{same} formulation as our GNCL objective in Eq. \ref{eq:GNCL1} with the cross entropy loss\footnote{This is not explicitly stated in the paper, but can be observed in the original implementation \url{https://github.com/Philip-Bachman/Pseudo-Ensembles}.}. We conclude, that the proposed bias-variance decomposition also encapsulates pseudo-ensembles and the GNCL objective can be viewed as an empirical version of this. However, we note that Pseudo-Ensembles have a very different viewpoint to our approach: Pseudo-Ensembles train a single network and spawn a diverse set of offsprings from this large network, whereas GNCL combines a set of smaller models into a large one.

\section{Experiments}
\label{sec:experiments}

In our experimental evaluation, we study two different aspects of the generalized bias-variance decomposition. As discussed before, when the base learners achieve zero training loss then the bias-variance decomposition implies, that the ensemble should not have any diversity. However, when applied to new, unseen data the base learners will likely have a nonzero loss and therefore, for a better generalization, it might be worthwhile to enforce some diversity during training. We will study this phenomenon in the context of Deep Learning by training ensembles with base learners of different capacities. For our evaluation, we use the CIFAR100 dataset \cite{Krizhevsky/2009} which contains $50000$ $32\times32\times 3$ images of various everyday objects which belong to one of $100$ classes. For testing, we utilize the given test split with $10000$ images. In all experiments, we perform standard data augmentation during training (Random cropping, random horizontal flipping, and normalization). We train for $100$ epochs with the AdaBeliefe \cite{zhuang/etal/2020} optimizer with a batch size of $128$ using PyTorch \cite{Paszke/etal/2019}. The initial learning rate is set to $0.001$ and halved every $25$ epochs. We evaluate ensembles utilizing three different types of base models: Low capacity, mid-capacity, and large-capacity ones. 

To do so, we use a ResNet architecture with $4$ residual blocks, an input convolutional, and a linear layer for the output. All convolutions have a kernel size of $3 \times 3$ with padding and stride of one. They are always followed by a BatchNorm layer and ReLu activation. Each residual block consists of two convolutions and the residual connection followed by a $2 \times 2$ max pooling. In total, each network has $9$ convolution layers and a single linear layer. To vary the model capacity we use a different number of filters in each base model. The mid-capacity model utilizes $32$ filters in each layer leading to $88~100$ trainable parameters. The large-capacity model utilizes $96$ filters leading to $706~468$ trainable parameters. As the low-capacity model, we use a binarized version of the mid-capacity model which constrains the weights and activations to $\{-1+1\}$. Binarized Neural Networks are a resource-friendly variation of `regular' floating-point Neural Networks that are optimized towards minimal memory consumption and fast model application. They have been shown to perform nearly as good as their floating-point siblings while being more resource efficient \cite{Hubara/etal/2016, Rastegari/etal/2016, Zhu/etal/2019, Buschjaeger/etal/2020a}. To train these models we use stochastic binarization, which retains the floating-point weights during the backward-pass but binarizes them during the forward pass as explained in \cite{Hubara/etal/2016}. Please note, that our binarized models have \textit{both}, weights and activations constrainted to $\{-1+1\}$. More details on the model architecture and training procedure can be found in our code at \url{https://github.com/sbuschjaeger/gncl}.

We also evaluated EfficientNet-B0 \cite{Tan/Le/2019}, MobilenetV3 \cite{Howard/etal/2019} and DenseNets \cite{Huang/etal/2017} as base learners which yielded similar performance with more parameters and longer training times. We note, that EfficientNet-B0 and MobilenetV3 are optimized towards the larger images of ImageNet \cite{deng/etal/2009} (typically $224 \times 224$ pixels) and heavily downsample input images in the first layers. We hypothesize that this is not really necessary on the comparably smaller images of CIFAR100 leading to a similar performance with more parameters. We include additional experiments on FashionMNIST, Imagenette and ImageNet in the appendix. 

As discussed previously, GNCL encapsulates many existing methods and thus we compare GNCL to those methods not directly captured by it. We compare ensembles with $M = 16$ models trained via Bagging, via Stochastic Multiple Choice Learning (SMCL), via Gradient Boosting (GB), via Snapshot Ensembing (SE) and with Generalized Negative Correlation Learning (GNCL) all minimizing the cross-entropy loss. For GNCL, we vary the regularization trade-off $\lambda \in \{0,0.1,0.2,\dots, 1.0\}$. Note, that GNCL with $\lambda = 0$ can be viewed as independent (Ind.) training of each network similar to Bagging but without bootstrap sampling. Similarly, for $\lambda = 1.0$ we train the ensemble in an End-To-End (E2E) fashion. For SE we take a snapshot of the model during optimization at the beginning of epochs 
$\{2,3,4,5,10,15,20,25,30,40,\dots, 90\}$ and combine them with the final model after $100$ epochs. Last, we also train a single model for reference. Please note, that these experiments are not meant to produce benchmarking result on the CIFAR100 dataset, but to investigate the effects of ensembling with different algorithms and base learners.

\begin{figure*}[t]
  \includegraphics[width=0.97\textwidth]{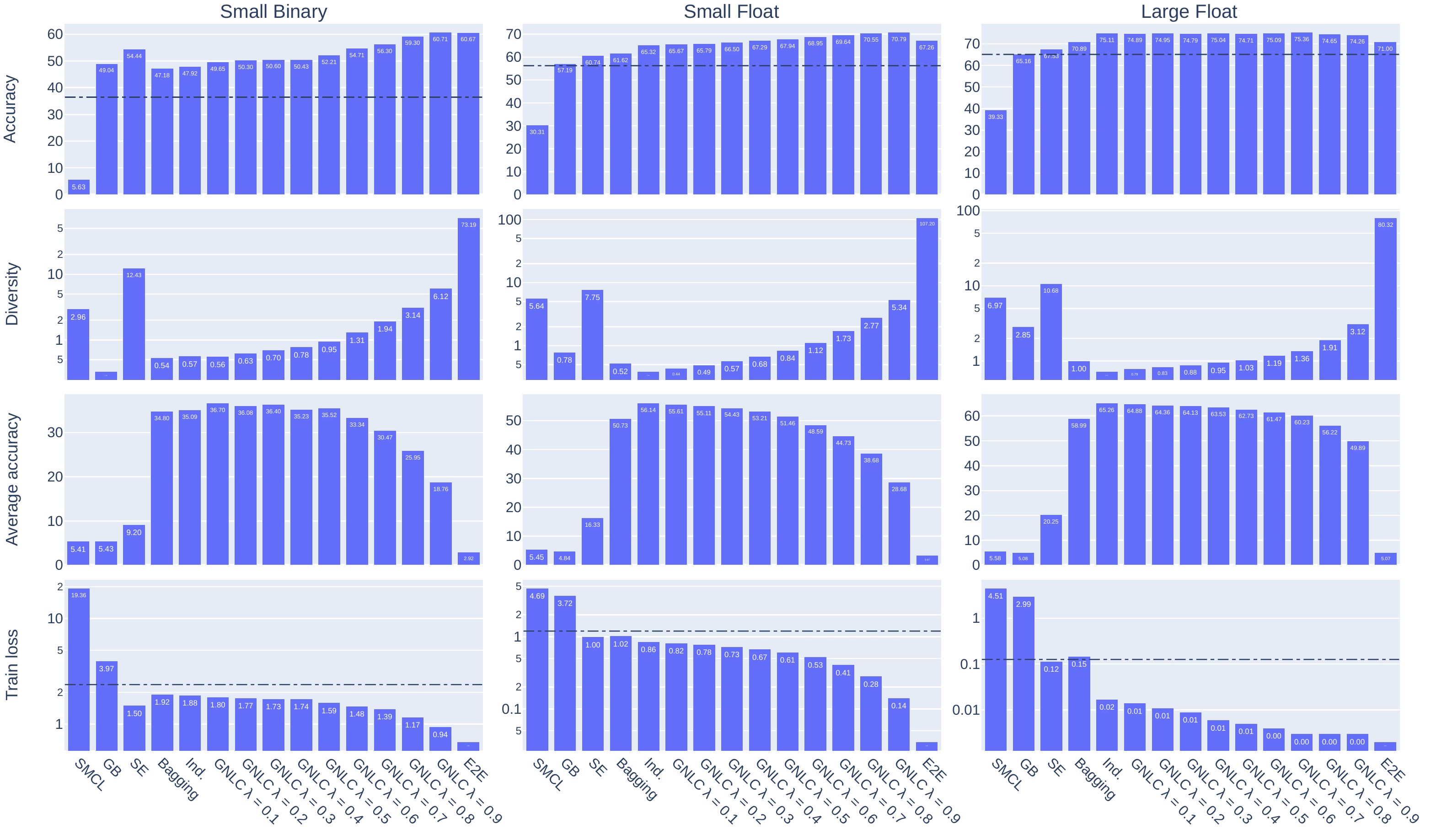}
  \caption{Experimental evaluation of various ensembling methods (E2E, Bagging, Gradient Boosting, GNCL with varying $\lambda$, SMCL and SE) with an ensemble size of $M = 16$ on CIFAR 100. Each column depicts a different base learner with low, mid, and large capacity (from left to right). The first row depicts the test accuracy of the ensemble, the second row shows the average diversity among the ensemble members (evaluated on test data), the third row shows the average accuracy of each expert and the last row depicts the training loss. The horizontal bar depicts the test accuracy and training loss of a single model.\label{fig:dl-experiments}}
\end{figure*}

Figure \ref{fig:dl-experiments} depicts the results of this experiment. Each column depicts a different base learner with low, mid, and large capacity (from left to right). The first row depicts the test accuracy of the ensemble, the second row shows the average diversity among the ensemble members (evaluated on the test set), the third row shows the average accuracy of each expert model and the last row depicts the training loss. The horizontal bar depicts the test accuracy and training loss of a single model. Looking at the low capacity binarized neural networks, we see that they achieve an accuracy of around $5 - 60 \%$. The clear winner in this setup is GNCL with $\lambda = 0.9$ and $\lambda = 1.0$ (E2E) achieving the highest accuracy, whereas SMCL is the worst with roughly $5 \%$ accuracy which is even below a single classifier. GNCL for smaller $\lambda$ behaves similar to Bagging and Gradient Boosting and with larger $\lambda$ there is a clear trend that the accuracy increases. Looking at the diversity we see that with larger $\lambda$ it steeply increases while the average test accuracy expectantly decreases. Interestingly, SMCL offers a similar diversity to GNCL with $\lambda = 0.8$, but with much worse (average) test accuracy. The same effect can be observed for SE, but much less sever. The training loss indicates that a single model is under parameterized for the task at hand achieving a loss in the range of $2$. Using more models increases the ensembles' capacity, and therefore decreases the overall loss. Expectantly, the test accuracy roughly follows the training loss: The smaller the loss, the better the test accuracy where again E2E is the best. Looking at the mid-capacity models we see a similar picture as before, but note that the optimal test accuracy now shifts towards a smaller $\lambda$ in the range of $0.8 - 0.9$. Again, the diversity increases with increasing $\lambda$ while the average test accuracy decreases. And again, we see that a smaller loss generally comes with better test accuracy. However, we note that while E2E learning offers by far the smallest loss it does \textit{not} achieve the best test accuracy. This effect becomes more extreme when looking at large-capacity models. Here, GNCL with smaller $\lambda$ in the range of $0 - 0.7$ seems to be best, whereas for larger $\lambda > 0.7$ the performance reduces. As expected, the diversity increases with increasing $\lambda$ while the average test accuracy decreases. Again we see, that E2E learning offers the smallest overall training loss, but it does \textit{not} achieve the best test accuracy. 

\textbf{We conclude:} For smaller capacity base models which do not achieve zero loss on their own, larger $\lambda$ values and E2E learning seems to be best. In these cases, a large diversity \textit{can} be beneficial as seen shown by the E2E approach. Once the base models become larger so that they achieve smaller losses on their own, enforcing diversity \textit{can} be hurtful. This is clearly shown by SMCL which produces very diverse ensembles with sub-optimal performance but also shown by E2E learning for mid and large-capacity base learners. 
In this case, the training should shift towards a more independent training of each model with a smaller $\lambda$ or independent training.

\section{Conclusion}
\label{sec:conclusion}

Ensemble learning plays a key role in many machine learning applications and offers state of the art performance. One of the guiding principles in designing an efficient ensembling algorithm is to enforce diversity in the ensemble. The theoretical roots of this approach lie in the bias-variance decomposition of the MSE loss which inspired many different approaches beyond the minimization of the MSE itself. While decompositions for other loss functions exist, they rarely inspired new learning algorithms beyond the general notion that diversity is important. In this paper, we studied the bias-variance decomposition for different loss functions more closely. We proposed a generalized bias-variance decomposition for twice differentiable loss functions which implies that the diversity depends on the covariance of the experts' outputs as well as the Hessian of the loss function. We derived a novel Generalized Negative Correlation Learning (GNCL) algorithm from it and detailed, how this algorithm encapsulates many existing works in literature. In an extensive experimental study we showed that diversity in the context of Deep Learning should not always be the main concern, but in fact, depends on the capacity of the base learners. For small-capacity base learners, diversity can be very beneficial as it allows the ensemble to minimize the overall loss more aggressively. For large-capacity base learners, diversity is also important but might be hurtful at a certain point as it artificially reduces the performance of the base learners and thus hurts their bias. This opens up the question, what Neural Network architectures are better suited for ensembling and how the optimization process impacts these results which we want to explore in the future.

\bibliography{literatur}
\bibliographystyle{icml2019}

\end{document}


\onecolumn
    \icmltitle{
        APPENDIX \\ A Generalized Bias-Variance Decomposition for Ensemble Learning
    }




\begin{icmlauthorlist}
\icmlauthor{Sebastian Buschjäger}{tu}
\icmlauthor{Lukas Pfahler}{tu}
\icmlauthor{Katharina Morik}{tu}
\end{icmlauthorlist}

\icmlaffiliation{tu}{Artificial Intelligence Group, TU Dortmund University, Germany}

\icmlcorrespondingauthor{Sebastian Buschjäger}{sebastian.buschjaeger@tu-dortmund.de}

\icmlkeywords{Bias, Variance, Ensemble, Deep Learning}

\vskip 0.3in



\printAffiliationsAndNotice{}  

\begin{abstract}
This is the appendix for our paper titled ``A Generalized Bias-Variance Decomposition for Ensemble Learning". This paper contains detailed derivations of all formulas in the paper which have been shortened for presentation. We also include additional experiments on more datasets. The sections here are meant to be a drop-in replacement of the corresponding sections in the paper. If a section is left empty, nothing changed compared to the paper. 
\end{abstract}

\section{Introduction}
\label{sec:introduction}
No changes. 

\section{Related Work}
\label{sec:related-work}
No changes. 

\section{A Generalized Bias-Variance Decomposition}
\label{sec:bias-variance}

We consider a supervised learning setting, in which we assume that training and test points are drawn i.i.d. according to some distribution $\mathcal D$ over the input space $\mathcal X$ and labels $\mathcal Y$. For training, we have given a labelled sample $\mathcal{S} = \{(x_i,y_i)|i=1,\dots,N\}$, where $x_i \in \mathcal X \subseteq \mathbb R^d$ is a $d$-dimensional feature-vector and $y_i\in \mathcal Y \subseteq \mathbb R^C$ is the corresponding target vector. For binary classification problems we set $C=1$ and $\mathcal Y = \{-1,+1\}$; for regression problems we have $C=1$ and $\mathcal Y = \mathbb R$. For multiclass problems with $C$ classes we encode each label as a one-hot vector $y = (0,\dots,0,1,0,\dots,0)$ which contains a `$1$' at coordinate $c$ for label $c \in \{0,\dots,C-1\}$. 

Given a model class $\mathcal H = \{\h \colon \mathcal X \to \mathbb R^C\}$ we wish to select that model that fits our current sample. In practice we employ various learning algorithms to do so, e.g. SGD in the context of Deep Learning or CART in the context of decision trees. These algorithms often introduce some form of randomization, e.g. by random initialization of weights or random sampling of splits thus introducing some model distribution $\Theta$. Yet, in the heart of these algorithms, we find the minimization of a loss function $\ell\colon \mathcal Y \times \mathcal Y \to \mathbb R_+$ which quantifies the error of our model's prediction $\h(x)$ compared to the real target $y$. So to choose the optimal algorithm for a problem, we may favor that algorithm which consistently produces the best models with the smallest loss: 
$$
\Theta^* = \arg\min_{\Theta}\E{h \sim \Theta,(x,y)\sim\mathcal D}{\ell(h(x),y)}
$$
For the rest of this paper we assume that $\ell$ is at least twice continuous differentiable and present appropriate choices at the end of this section. We now use a second-order Taylor approximation of $\ell$ around the centre $\m(x) = \E{h\sim \Theta}{h(x)}$. For readability we now drop the subscript $h \sim \Theta,(x,y)\sim\mathcal D$. Similarly we write $h(x) = h$ and $\m(x) = \m$ and $\ell(h(x),y) = \ell(h)$:
\begin{align*}
    \E{}{\ell(h)}
    &= 
    \E{}{\ell(\mu)} +
    \E{}{(\h - \m)^T\nabla_{\m} \ell(\m)} + \E{}{\frac{1}{2} (\h-\m)^T \nabla^2_{\m}\ell(\m) (\h-\m)} + \E{}{R_3} 
\end{align*}
where $R_3$ denotes the remainder of the Taylor approximation containing the third and higher derivatives.

We note, that $\nabla_{\m} \ell(\m)$ does not depend on $\h$ since $\m$ is a constant given a fixed test point $\x$ and therefore $\E{}{\nabla_{\m}  \ell(\m)} = \nabla_{\m} \ell(\m)$. Also note, that per definition $\E{}{h} = \m$ so that the second summand vanishes: 
\begin{align*}
\E{\h}{(\h - \m)^T\nabla_{\m} \ell(\m))} 
    &= \E{\h}{(\h - \m)^T}\nabla_{\m} \ell(\m)  \\
    &= \left(\E{}{\h} - \E{}{\m} \right)^T \nabla_{\m} \ell(\m)  \\
    &= \left(\m - \m \right)^T \nabla_{\m} \ell(\m)  \\
    &= 0
\end{align*}

Naturally the quality of this approximation depends on the magnitude of the remainder and it becomes exact if the loss function does not have a third derivative. Otherwise a classic text-book (see e.g. \cite{edwards/2012, koenigsberger/2013}) result bounds the magnitude of the remainder for functions which are $k+1$ times continuous differentiable. Let $M, r > 0$ be constants with 
\begin{align*}
    \norm{\h - \m}{1} &< r \\
    |D^{\alpha} \ell(\m,y)| &\le M \frac{s!}{r^s} \text{~~for~} |\alpha| = s
\end{align*}
where we used the multiindex notation $\alpha = (\alpha_1,\dots,\alpha_n)\in \mathbb N_0^n$ with $|\alpha| = \alpha_1+\dots+\alpha_n$ and $D^{\alpha} f = D_1^{\alpha_1}f...D_n^{\alpha_n}f$. Then every Taylor Series with $|\h - \m|_1 \le \rho < r$ converges with 
\begin{align*}
    R_s\cdot \norm{\h - \m}{1} \le M\left(\frac{\rho}{r}\right)^{s+1} 
\end{align*}

To be more useful we reformulate this expression. Let there be some $m\in\mathbb R$ so that $|D^{\alpha} \ell(\m, y)| \le m$ for all $y,\m$ then
$$
m = M \frac{s!}{r^s} \Rightarrow M = \frac{1}{s!} m r^s
$$
Re-substituting
\begin{align*}
    R_s\cdot \norm{\h - \m}{1} &\le M\left(\frac{\rho}{r}\right)^{s+1} = \frac{1}{s!} m r^s \left(\frac{\rho}{r}\right)^{s+1} \\
    &= \frac{1}{s!} m r^s \frac{\rho^{s+1}}{r^{s+1}} = \frac{1}{s!} m \frac{\rho}{r} \rho^s \\
    &\le \frac{1}{s!} m \rho^s \le \frac{1}{s!} m \max_{\h}\norm{\h - \m}{1}^s \\
    &= \frac{1}{s!} m \sum_{i=1}^{C} \max_{h_i} (h_i - \mu_i)^s \\
    &\le \frac{1}{s!} m C \max_{h_1,\dots,h_{C}} (h_i - \mu_i)^s
\end{align*}

where the last line holds due to $\frac{\rho}{r} \le 1$ since $\rho < r = \max_{\h}\norm{\h-\m}{1}$. By setting $s=3$ we find the remainder for the second order Taylor approximation:
$$
R_3(\h - \m) \le \frac{1}{6}m \max_{\h}\norm{\h - \m|}{1}^3 \le \frac{1}{6} m C \max_{h_1,\dots,h_{C}} (h_i - \mu_i)^3
$$

For a sufficiently small remainder we approximate:
\begin{align}
    \label{eq:covar-decomposition}
    \E{}{\ell(h)} 
        &\approx \E{}{\ell(\mu)} + \E{}{\frac{1}{2} \phi^T \nabla^2_{\m}\ell(\m) \phi} \\ 
        &= \E{}{\ell(\mu)} + \frac{1}{2} \E{}{(\h-\m)^T} \nabla_{\m} \ell(\m) \E{}{(\h-\m)^T} + \frac{1}{2}tr\left(\nabla_{\m} \ell(\m,y) \texttt{cov}(\h-\m, \h-\m)\right)  \\ 
        &= \E{}{\ell(\mu)} + \frac{1}{2}tr\left(\nabla_{\m} \ell(\m) \texttt{cov}( \phi,  \phi)\right) 
\end{align}
where $\phi = (\h-\m)$ and the second line is the quadratic form of the expectation. We may interpret this decomposition as a generalized Bias-(Co-)Variance decomposition: While the LHS depicts the expected error of a model $h$, the first term on the RHS depicts the error of the expected model - or differently coined the \textit{algorithm's} bias. Moreover, the second term can be interpreted as the co-variance of $h$ with respect to the expected model $\m$ given a loss-specific multiplicative constant $\nabla^2 \ell(\mu)$.  

\subsection{Example 1: Mean-squared error}
Consider the mean squared error (MSE) of a one dimensional regression task $\mathcal Y = \mathbb R$ and let $z = h(x)$:
\begin{align*}
    \ell(z,y) &= \frac{1}{2}(z - y)^2 \\
    \frac{\partial \ell}{\partial z} &= (z - y) \\
    \frac{\partial^2 \ell}{\partial z \partial h(x)} &= 1 \\
    \frac{\partial^3 \ell}{\partial z \partial h(x) \partial z} &= 0 
\end{align*}
The third derivative of the MSE vanishes and thus the above approximation is exact. The resulting decomposition matches exactly the well-known Bias-Co-Variance decomposition.

\subsection{Example 2: Negative-likelihood Loss}
As a second example we consider multi-class classification problem with $C$ classes. Let $z = h(x) \in \mathbb R^C$ and let $\ell$ be the negative-likelihood loss (NLL):
\begin{align*}
    \ell(z,y) &= -\sum_{i=1}^C y_i \log(z_i) \\
    \frac{\partial \ell}{\partial z_i} &= -\frac{y_i}{z_i} \\
    \frac{\partial^2 \ell}{\partial z_i \partial z_j} &= -\frac{y_i}{z_i^2} (-1) \mathbbm 1\{i=j\} = \frac{y_i}{z_i^2}\mathbbm 1\{i=j\} \\
    \frac{\partial^3 \ell}{\partial z_i \partial z_j \partial z_k} &= -2\frac{y_i}{z_i^3}\mathbbm 1\{i=j=k\} 
\end{align*}
For this loss function, the third derivative does not vanish and thus the decomposition is not exact. Looking at the third derivative we also see, that it can get uncontrollably large for $z_i \to 0$ if $y_i = 1$. Thus, if a model completely fails with a wrong prediction then the decomposition error can be unbounded. Put differently, the performance of a model using the NLLLoss cannot be completely explained in terms of `Bias' and `Variance' since the remainder is not neglectable. 

\subsection{Example 3: Cross Entropy Loss}

A common combination in Deep Learning is the NLLLoss with the softmax function. Again, let $z = h(x) \in \mathbb R^C$. The softmax function maps each output dimension $z_i$ of the classifier to a probability:
$$
q_i = \frac{e^{z_i}}{\sum_{i=1}^C e^{z_j}}
$$
We combine softmax with the NLLLoss:
\begin{align*}
    \ell(z,y) &= -\sum_{i=1}^C y_i \log(z_i) = -\sum_{i=1}^C y_i \log \left(\frac{e^{z_i}}{\sum_{i=1}^C e^{z_j}} \right) \\
    \frac{\partial \ell}{\partial z_i} &= q_i - \mathbbm 1\{y_i = 1\} \\
    \frac{\partial^2 \ell}{\partial z_i \partial z_j} &= q_i \left( \mathbbm 1\{i=j\} -q_j\right) = \begin{cases} q_i (\left( 1 -q_j\right)) \quad i==j \\ -q_i q_j \quad \quad \quad \quad\text{else}\end{cases} \\
    \frac{\partial^3 \ell}{\partial z_i \partial z_j \partial z_k} &= \underbrace{\mathbbm 1\{i=j\}q_i \left( \mathbbm 1\{i=k\} - q_k \right)}_{\in [-1,1]}
    - \underbrace{q_i q_j\left( \mathbbm 1\{i=k\} - q_k \right)}_{[-1,1]} - \underbrace{q_i q_j \left( \mathbbm 1\{j=k\} - q_k \right)}_{[-1,1]}
\end{align*}
Due to the softmax function we have $\sum_{c=1}^C q_c = 1, q_c > 0~\forall c = 1,\dots,C$. The maximum of the third derivative is obtained for pairwise unequal $i,j,k$ ($i \not= j, j \not= k, i \not= k$) and $q_i = q_j = q_k = \frac{1}{3}$:
$$
2 \cdot q_i q_j q_k \le \frac{1}{27} < 0.038
$$
Thus, the decomposition error for the cross entropy loss is bounded and we can explain a models performance in terms of its Bias and Variance (up to the bounded remainder).  

\subsection{Example 4: Exponential loss}
The exponential loss can also be used for (binary) classification problem with $\mathcal Y = \{-1,+1\}$ and $C=1$:
\begin{align*}
    \ell(h(x),y) &= \exp(-h(x) y) \\
    \frac{\partial \ell}{\partial h(x)} &= -y \exp(-h(x) y) \\
    \frac{\partial^2 \ell}{\partial h(x) \partial h(x)} &= y^2\exp(-h(x) y) = \exp(-h(x) y) \\
    \frac{\partial^3 \ell}{\partial h(x) \partial h(x) \partial h(x)} &= -y\exp(-h(x) y) 
\end{align*}
For this loss function, the third derivative does not vanish and thus the decomposition is not exact. We may estimate the remainder. Let $h(x)\in [-1,+1]$, then  
$$
\frac{\partial^3 \ell}{\partial h(x) \partial h(x) \partial h(x)} \le \exp(1)
$$
and therefore
$$
R_3(\h - \m) \le \frac{1}{6} \exp(1) \le 0.454
$$
Note that if we predict one example wrong we already suffer a loss of $\exp(1)\approx 2.7$ which easily dominates the remainder.

\subsection{Example 5: Gaussian Hinge Loss}
Last, we present a variant of the popular hinge loss function. Since the normal hinge function is not differentiable and variants like smooth hinge and squared hinge do not have smooth second derivatives, we consider a continously differentiably variant based on the Gaussian error function for a binary classification problem with $\mathcal Y = \{-1,+1\}$ and $C=1$:
\begin{align*}
    \ell(h(x),y) &= \frac{e^{-h(x)^2}}{\sqrt\pi} - yh(x) [1 +\mathrm{erf}(-yh(x))] \\
    \frac{\partial \ell}{\partial h(x)} &= -y \exp(-h(x) y) \\
    \frac{\partial^2 \ell}{\partial h(x) \partial h(x)} &= y^2\exp(-h(x) y) = \exp(-h(x) y) \\
    \frac{\partial^3 \ell}{\partial h(x) \partial h(x) \partial h(x)} &= -y\exp(-h(x) y) 
\end{align*}
 
Similar to the exponential loss function, the third derivative does not vanish and thus the decomposition is not exact. We may estimate the remainder. Let $h(x)\in [-1,+1]$, then  
$$
\frac{\partial^3 \ell}{\partial h(x) \partial h(x) \partial h(x)} = -y\exp(-h(x) y) \le \exp(1)
$$
and therefore
$$
R_3(\h - \m) \le \frac{1}{6} \exp(1) \le 0.454
$$
similar to the exponential loss.

\section{Generalized Negative Correlation Learning}
\label{sec:gncl}
No changes.

\section{Experiments}
\label{sec:experiments}
No changes.

\section{Additional Experiments}
\label{sec:experiments}

In addition to the CIFAR100 experiments presented in the paper we repeated the same experiments on additional datasets. The goal of these experiments is to offer a broader view of the impact of the model capacity for different tasks. 

\subsection{FashionMNIST}
In this experiment we use the FashionMNIST dataset \cite{xiao/2017} which is a variation of the well-known MNIST data-set. FashionMNIST contains $60000$ $28\times28$ greyscale images of various different clothing items belonging to $10$ classes. In all experiments, we perform limited data augmentation (random horizontal flipping) and no normalization during training. We train for $150$ epochs with the AdaBeliefe \cite{zhuang/etal/2020} optimizer with a batch size of $256$ using PyTorch \cite{Paszke/etal/2019}. The initial learning rate is set to $0.01$ and halved every $25$ epochs. We evaluate ensembles utilizing four different types of base models: Low capacity, low-mid-capacity, mid-large-capacity and large-capacity ones. 

To do so, we use a ResNet architecture with $4$ residual blocks, an input convolutional, and a linear layer for the output. All convolutions have a kernel size of $3 \times 3$ with padding and stride of one. They are always followed by a BatchNorm layer and ReLu activation. Each residual block consists of two convolutions and the residual connection followed by a $2 \times 2$ max pooling. In total each network has $9$ convolution layers and a single linear layer. To vary the model capacity we use two version of this architecture. The small version uses $32$ filters in each layer leading to $88~100$ trainable parameters and the large version uses  $96$ filters leading to $706~468$ trainable parameters. In addition, we use a binarized version of each architecture that constrains the weights and activations to $\{-1+1\}$. Binarized Neural Networks are a resource-friendly variation of `regular' floating-point Neural Networks that are optimized towards minimal memory consumption and fast model application. To train these models we use stochastic binarization, which retains the floating-point weights during the backward-pass but binarizes them during the forward pass as explained in \cite{Hubara/etal/2016}. 

Again, we compare ensembles with $M = 16$ models trained via Bagging, via Stochastic Multiple Choice Learning (SMCL), via Gradient Boosting (GB), via Snapshot Ensembing (SE) and with Generalized Negative Correlation Learning (GNCL) all minimizing the cross-entropy loss. For GNCL, we vary the regularization trade-off $\lambda \in \{0,0.1,0.2,\dots, 1.0\}$. Note, that GNCL with $\lambda = 0$ can be viewed as independent (Ind.) training of each network similar to Bagging but without bootstrap sampling. Similarly, for $\lambda = 1.0$ we train the ensemble in an End-To-End (E2E) fashion. For SE we take a snapshot of the model during optimization at the beginning of epochs 
$\{2,3,4,5,10,15,20,25,30,40,\dots, 90\}$ and combine them with the final model after $150$ epochs. Last, we also train a single model for reference. 

\subsection{Imagenette} 

In this experiment, we use the Imagentte dataset\footnote{\url{https://github.com/fastai/imagenette}} which contains a subset of the 9469 images from the ImageNet dataset belonging to 10 classes. Imagentte is meant as a drop-in replacement of ImageNet with comparable difficulty, but with much fewer images and classes to facilitate faster model development. We use the small $160$px version and downscale each image to the dimensions $128\times128\times3$. We apply standard normalization, as well as random cropping and random rotation during training. We train for $150$ epochs with the AdaBeliefe \cite{zhuang/etal/2020} optimizer with a batch size of $256$ using PyTorch \cite{Paszke/etal/2019}. The initial learning rate is set to $0.01$ and halved every $25$ epochs. We evaluate ensembles utilizing three different types of base models: Low-capacity, mid-capacity and large-capacity ones. 

To do so, we use a tiny version of the MobileNetV3 architecture \cite{Howard/etal/2019} which is derived from the small MobileNetV3 architecture by removing the last three bottleneck layers and reducing the number of filter to $[16, 72, 88, 96, 120, 120, 60, 72]$ for the respective bottleneck layers. The last linear layer has a size of $512$ and the entire architecture has $131~586$ trainable parameters. 

Again, we compare ensembles with $M = 16$ models trained via Bagging, via Stochastic Multiple Choice Learning (SMCL), via Gradient Boosting (GB), via Snapshot Ensembling (SE), and with Generalized Negative Correlation Learning (GNCL) all minimizing the cross-entropy loss. For GNCL, we vary the regularization trade-off $\lambda \in \{0,0.1,0.2,\dots, 1.0\}$. Note, that GNCL with $\lambda = 0$ can be viewed as independent (Ind.) training of each network similar to Bagging but without bootstrap sampling. Similarly, for $\lambda = 1.0$ we train the ensemble in an End-To-End (E2E) fashion. For SE we take a snapshot of the model during optimization at the beginning of epochs 
$\{2,3,4,5,10,15,20,25,30,40,\dots, 90\}$ and combine them with the final model after $150$ epochs. Last, we also train a single model for reference. 

\begin{figure*}
  \includegraphics[width=\textwidth]{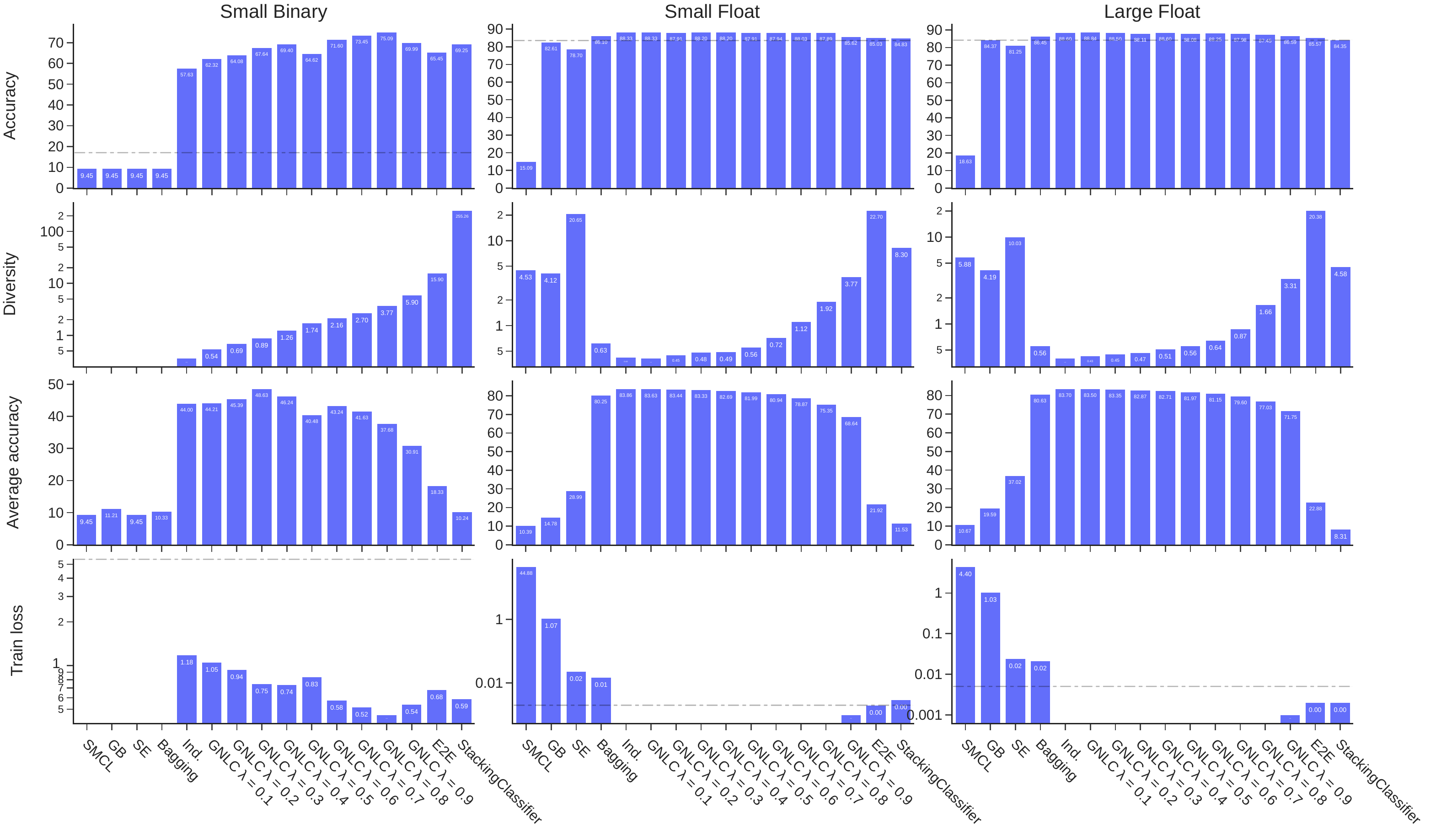}
  \caption{Experimental evaluation of various ensembling methods (E2E, Bagging, Gradient Boosting, GNCL with varying $\lambda$, SMCL and SE) with an ensemble size of $M = 16$ on Imagenette. Each column depicts a different base learner with low, mid, and large capacity (from left to right). The first row depicts the test accuracy of the ensemble, the second row shows the average diversity among the ensemble members (evaluated on test data), the third row shows the average accuracy of each expert and the last row depicts the training loss. The horizontal bar depicts the test accuracy and training loss of a single model.\label{fig:imagenette}}
\end{figure*}

Figure \ref{fig:imagenette} depicts the results. Each column depicts a different base learner with low, mid, and large capacity (from left to right). The first row depicts the test accuracy of the ensemble, the second row shows the average diversity among the ensemble members (evaluated on the test set), the third row shows the average accuracy of each expert model and the last row depicts the training loss. The horizontal bar depicts the test accuracy and training loss of a single model. Looking at the low capacity binarized neural networks, we see that they achieve an accuracy of around $10 - 75 \%$. The clear winner in this setup GNCL with $\lambda = 0.8$ achieving the highest accuracy, whereas SMCL, GB, SE, and Bagging are the worst with roughly $10 \%$ accuracy which is even below a single classifier. After further examination, we found that these models would not learn at all in this setup, possibly due to the large step-sizes used.  Looking at the diversity we see that with larger $\lambda$ it steeply increases while the average test accuracy behaves somewhat chaotic. The training loss indicates that a single model is under parameterized for the task at hand achieving a loss in the range of $5$. Using more models increases the ensembles' capacity, and therefore decreases the overall loss. Expectantly, the test accuracy roughly follows the training loss: The smaller the loss, the better the test accuracy. Looking at the mid-capacity models we see a similar picture as before but note that the optimal test accuracy now shifts towards a smaller $\lambda$. Here, the algorithms all behave similarly for $\lambda$ in the range of $0.1 - 0.8$. Again, the diversity increases with increasing $\lambda$ while the average test accuracy slowly decreases. And again, we see that a smaller loss generally comes with better test accuracy. Looking at the high-capacity models we see roughly the same picture which can be expected since the mid-capacity base learners already achieved a loss close to zero. 

\subsection{ImageNet}

\section{Conclusion}
\label{sec:conclusion}
No changes. 

\bibliography{literatur}
\bibliographystyle{icml2019}